\documentclass[11pt,a4paper]{article}
\usepackage[hyperref]{acl2020}
\usepackage[utf8]{vietnam}
\usepackage[utf8]{inputenc}
\usepackage[vietnamese=nohyphenation]{hyphsubst}
\usepackage{times}
\usepackage{latexsym}
\usepackage{graphicx}
\usepackage{multirow}

\usepackage{microtype}

\aclfinalcopy 

\title{ComOM at VLSP 2023: A Dual-Stage Framework with BERTology and Unified Multi-Task Instruction Tuning Model for Vietnamese Comparative Opinion Mining}

\author{Dang Van Thin, Duong Ngoc Hao, Ngan Luu-Thuy Nguyen \\
  University of Information Technology-VNUHCM \\
  Vietnam National University, Ho Chi Minh City, Vietnam \\
  \texttt{\{thindv,haodn,ngannlt\}@uit.edu.vn}
}

\date{}

\begin{document}
\maketitle
\begin{abstract}
The ComOM shared task aims to extract comparative opinions from product reviews in Vietnamese language. There are two sub-tasks, including (1) Comparative Sentence Identification (CSI) and (2)  Comparative Element Extraction (CEE). The first task is to identify whether the input is a comparative review, and the purpose of the second task is to extract the quintuplets mentioned in the comparative review. To address this task, our team proposes a two-stage system based on fine-tuning a BERTology model for the CSI task and unified multi-task instruction tuning for the CEE task. Besides, we apply the simple data augmentation technique to increase the size of the dataset for training our model in the second stage. Experimental results show that our approach outperforms the other competitors and has achieved the top score on the official private test.
\end{abstract}

\section{Introduction}
\label{introduction}
Recently, the comparative opinion mining or Comparative Opinion Quintuple Extraction (COQE) has received considerable attention by research community, especially for low-resource languages \cite{varathan2017comparative}. The purpose of this task at the shared-task challenge is aim to identify comparative reviews and extract the pre-defined opinon quintuples, i.e (subject, object, aspect, predicate, and comparison label) \cite{jindal2006mining}. For example, given a review ``\textit{iPhone 14 Pro Max has a better battery life compared to its competitors}'', the comparative quintuple is extracted as \{"\textit{subject}": ["1\&\&iPhone", "2\&\&14", "3\&\&Pro", "4\&\&Max"], "\textit{object}": ["12\&\&its", "13\&\&competitors"], "\textit{aspect}": ["8\&\&battery", "9\&\&life"], "\textit{predicate}": ["7\&\&better"], "\textit{label}": "COM+"\}. Different from previous studies \cite{jindal2006mining,liu-etal-2021-comparative}, this task requires the competitors to extract the text information and its order instead of only the text from the review. Moreover, it is noted that a review can contain several associated comparative quintuples. 

In order to tackle the comparative opinion mining task, \cite{liu-etal-2021-comparative} presented a multi-stage neural network framework based on the pre-trained BERT models \cite{devlin-etal-2019-bert} to address three single sub-task and then combine the results for final comparative quintuples. Instead of solving the task as multiple stages, \cite{10191436} proposed an end-to-end model by utilizing the BERT \cite{devlin-etal-2019-bert} and Graph Convolutional Network (GCN) to represent the input feature. Then, a non-autoregressive decoder is applied to predict all the quintuples mentioned in the input. Similarly, the work of \cite{yang-etal-2023-unicoqe} introduced a set-matching strategy for the training paradigm with a unified generative model for the COQE task. This approach helps their model overcome the error propagation problem of previous systems \cite{liu-etal-2021-comparative}.

In this paper, we present our solution for the Comparative Opinion Mining from Vietnamese Product Reviews challenge at VLSP 2023 \cite{hoang2023overview}. To tackle this task, we proposed a two-stage framework, which consists of: (1) the first stage, which identifies comparative sentences, and (2) the second stage, which extracts five comparative elements, referred to as the list of opinion quintuples. In the first stage, we apply a binary classification model based on the pre-trained language model PhoBERT \cite{nguyen-tuan-nguyen-2020-phobert} to identify the comparative review. In the second stage, we present an end-to-end multi-task instruction tuning model to predict the quintuples. Besides, we apply the simple data augmentation technique to increase the size of the dataset for training our model in the second stage. Experimental results show that our approach outperforms the other competitors and has achieved the top score on the official private test\footnote{https://aihub.ml/competitions/601\#results}. 

The remainder of this paper is structured as follows: In Section \ref{method}, we present our end-to-end architecture for the COQE task in Vietnamese reviews. The details of our experiments are presented in Section \ref{experiment}, including the experimental settings, baseline models, and the main results. Finally, in Section \ref{conclusion}, we summarize the conclusions of this paper and discuss potential directions for future research.

\section{Methodology}
\label{method}

\subsection{Task Formulation}
First, the formulation of the COQE task is presented as follows: Given a review either at the sentence-level or document-level denoted as $X = \{x_1, x_2, ...., x_n\}$, where \textit{n} represents the number of words, the primary objective of this task is to determine whether a given sentence is comparative and, if it is, to extract all $m$ comparative quintuples mentioned in the review: 
\begin{equation}
    Y = \{quin_1; quin_2; ...; quin_m\}
\end{equation}
with $quin = \{sub, obj, asp, pred, label\}$ is an extracted quintuple, where $sub$ denotes the subject of the comparison, $obj$ refers to the object being compared to the subject, $asp$ is the word or phrase describing the feature or attribute of both subject and object that is under comparison, $pred$ is the comparative word or phrase used to express the comparison, $label$ represents the comparison type label, which can take one of eight possible values: EQL, DIF, COM, COM+, COM-, SUP, SUP+, SUP-. Note that the first four elements of the quintuple must be extracted along with their corresponding index positions from the given review.  Additionally, only the predicate is mandatory in each quintuple; the remaining three elements among the first four are optional and may be omitted.

Based on the task formulation above, there are two sub-tasks to complete: (1) Comparative Sentence Identification (CSI) - This involves identifying whether a given review is comparative or not; (2) Comparative Element Extraction (CEE) - This task is aimed to extract tuples of five comparative elements from the given review.

\subsection{Stage 1: Comparative Sentence Identification}
To identify whether the sentence belongs to the comparative review or not, we implement a binary classification model based on fine-tuning encoder-type BERT pre-trained language models. Specifically, given an input review $X = \{x_1, x_2, ...., x_n\}$ with n tokens, we feed it the BERT modelS to obtain the hidden representations $H = \{h_{cls},h_1, ...., x_n, h_{sep}\}$ after adding the special tokens (CLS and SEP) in the last layer. Then, we extract the hidden state $h_{cls}$ of CLS token as the final review representation to a classification layer to compute the probability:
\begin{equation}
    \hat{y} = softmax(W\cdot h_{cls} + b)
\end{equation}
where $W$ and $b$ are the weight matrix and bias term for the classification layer.

\subsection{Stage 2: Comparative Element Extraction}

\begin{table*}[t]
\centering
\caption{The list of instruction templates for all sub-tasks with corresponding outputs for the example input.}
\label{mutktas}
\resizebox{\textwidth}{!}{%
\begin{tabular}{ll}
\hline
\multicolumn{2}{l}{\textbf{Input:} iPhone 14 Pro Max has a better battery life compared to its competitors} \\ \hline
\multicolumn{1}{l|}{\textbf{Instruction}} & \textbf{Output} \\ \hline
\multicolumn{1}{l|}{\begin{tabular}[c]{@{}l@{}}Please extract five elements including subject, object, aspect, predicate, and comparison type in the sentence\end{tabular}} & (iPhone 14 Pro Max; its competitors; battery life; better; COM+) \\ \hline

\multicolumn{1}{l|}{\begin{tabular}[c]{@{}l@{}}Please extract four elements including subject, object, aspect, and predicate in the sentence\end{tabular}} & (iPhone 14 Pro Max; its competitors; battery life; better) \\ \hline

\multicolumn{1}{l|}{\begin{tabular}[c]{@{}l@{}}Please extract four elements including subject, object, aspect, and comparison type in the sentence\end{tabular}} & (iPhone 14 Pro Max; its competitors; battery life; COM+) \\ \hline

\multicolumn{1}{l|}{\begin{tabular}[c]{@{}l@{}}Please extract three elements including subject, object, and aspect in the sentence\end{tabular}} & (iPhone 14 Pro Max; its competitors; battery life) \\ \hline

\multicolumn{1}{l|}{\begin{tabular}[c]{@{}l@{}}Please extract three elements including subject, object, and comparison type in the sentence\end{tabular}} & (iPhone 14 Pro Max; its competitors; COM+) \\ \hline

\multicolumn{1}{l|}{\begin{tabular}[c]{@{}l@{}}Please extract three elements including subject, object, and predicate in the sentence\end{tabular}} & (iPhone 14 Pro Max; its competitors; better) \\ \hline

\multicolumn{1}{l|}{\begin{tabular}[c]{@{}l@{}}Please extract three elements including aspect, predicate, and comparison type in the sentence\end{tabular}} & (battery life; better; COM+) \\ \hline

\multicolumn{1}{l|}{\begin{tabular}[c]{@{}l@{}}Please extract two elements including subject and object in the sentence\end{tabular}} & (iPhone 14 Pro Max; its competitors) \\ \hline

\multicolumn{1}{l|}{\begin{tabular}[c]{@{}l@{}}Please extract two elements including aspect and predicate in the sentence\end{tabular}} & (battery life; better) \\ \hline

\multicolumn{1}{l|}{\begin{tabular}[c]{@{}l@{}}Please extract two elements including aspect and comparison type in the sentence\end{tabular}} & (battery life; COM+) \\ \hline

\end{tabular}%
}
\end{table*}

In this stage, we present our solutions to extract all comparative elements in the quintuplets. To tackle this sub-task, we design generative paradigms combined with a multi-task instruction tuning strategy based on the encoder-decoder architecture. Specifically, we transform different CEE tasks into natural language generation problems and design an effective generation paradigm. Additionally, we introduce descriptive prompts to improve the understanding of generative models for each specific task in Vietnamese language. To represent the comparative quintuples as a generation template, we investigate two methods as the following example:
\begin{quote}
    \textbf{Target}: \{"\textit{subject}": ["1\&\&iPhone", "2\&\&14", "3\&\&Pro", "4\&\&Max"], "\textit{object}": ["12\&\&its", "13\&\&competitors"], "\textit{aspect}": ["8\&\&battery", "9\&\&life"], "\textit{predicate}": ["7\&\&better"], "\textit{label}": "COM+"\} \\\\
    \textbf{Template 1}: [s] \textit{Iphone 14 Pro Max} [/s] [o] \textit{its competitors} [/o] [a] \textit{battery life} [/a] [p] \textit{better} [/p] [l] \textit{COM+} [/l]  \\\\
    \textbf{Template 2}: \{\textit{Iphone 14 Pro Max}; \textit{its competitors}; \textit{battery life}; \textit{better}; \textit{COM+}\}\\
\end{quote}

In case the value of any element is empty, it will be represented as None. Note that the output template does not contain the order of elements (e.g. subject, object, aspect, predicate) in the input review. If there are many quintuplets assigned for an input, they are concatenated with ``;'' punctuation as the output of the model. To get the order of extracted information, we develop a fuzzy string-matching algorithm combined with heuristic rules (e.g. if there are two indexes, we will select the index with the shortest distance compared to other values). We hypothesise that there is a correlation information between sub-tasks and comparative elements in the quintuplets. Therefore, fine-tuning the generative models on the instruction with multi-task learning can improve the overall performance. Table \ref{mutktas} shows an example with the corresponding instructions and output for sub-tasks based on the combination of five elements. During the development phase, the second template yields better performance than the first template. Therefore, we use template 2 for our submission and experimental results. For the generative models, we fine-tune the two pre-trained T5 models, including the multilingual T5 \cite{mt5} and monolingual viT5 \cite{vit5}.

\subsection{Data Augmentation}
In this section, we present our augmentation strategy to obtain the diverse augmented training data. The idea of this method appeared when we analysed the prediction of models on the public test set. We find that the model can predict incorrectly when replacing any element in a comparative quintuplet that appears in the training set. Therefore, we introduce a simple data augmentation to diversify the training set. Our strategy is designed as follows: (1) First, we collect all values to create the specific word set for each element in the quintuple. (2) Subsequently, for each comparative review, we randomly replace the subject, object, and aspect from the word set to create the new sentence and corresponding quintuples. For the predicate and comparison labels, we randomly exchange them together because the comparison labels depend on the predicate value. (3) Finally, we balance the augmented samples based on the number of comparative elements and filter the duplicate samples and missing orders between quintuplets and review. We combine the original data combined with augmented data for training models.

\section{Experiments}
\label{experiment}
\subsection{Experimental Setup}

\textbf{Data and Preprocessing}: During the development phase, we used the official training set to train models and tune hyper-parameters on the public test set. However, we detected instances of mis-annotation within comparative reviews. As a result, we introduced a straightforward filtering component designed to exclude sentences that display a comparative structure but lack annotated quintuples. To achieve this, we employed a similarity technique based on sentence embedding\footnote{https://huggingface.co/sentence-transformers/LaBSE} and set a specific threshold (in our case, 80\%) to filter out such sentences. This approach enhances the performance of our binary classification model, reducing ambiguity in the Comparative Sentiment Identification task. For the final submission on the private set, we combined the training set with the public test set to train the model, following the recommendation of the organizing committee. In our system, we do not apply any specific pre-processing steps, except for removing multiple spaces. 
\newline\newline\textbf{Evaluation Metrics}: These are 120 metrics are evaluated based on the combination of comparative elements. However, in our paper, we report the macro and micro F1-score on exact match for entire quintuples. While the final ranking score in the scoreboard is determined by the exact match macro F1-score. \newline\newline\textbf{Configuration Settings}: We implemented our models using Trainer API from Hugging Face library \citep{wolf-etal-2020-transformers}. For the first stage, we experimented with different encoder-type language models, such as  PhoBERT \citep{nguyen-tuan-nguyen-2020-phobert}, XLM-R \cite{xlm-r}, etc. to identify the comparative review. To train the models, we set the maximum input length as 156 tokens, and the learning rate is set to 5e-5 with a batch size of 32. 

For the second stage, we utilize the T5 variants, including the viT5 \cite{vit5} and mT5 \cite{mt5} as the backbone model. The maximum input and output length is set as 156 tokens, and the number of epochs is set as 20 with a batch size of 16. We used an AdamW optimizer with a linear schedule warmup technique. The learning rate is set to 2e-5 for viT5 models and 3e-4 for mT5 models, respectively. We train our models on NVIDIA A100 80G GPUs.

\subsection{Comparative Models}
In addition to comparing the performance of our best model, we also employed the following baseline systems for each stage as below. For the  CSI task, we fine-tune different pre-trained language models, including the following multilingual and monolingual models.
\begin{itemize}
    \item \textbf{mBERT}: mBERT \cite{devlin-etal-2019-bert} is a pre-trained language model designed to handle text classification tasks and other NLP tasks across multiple languages. We fine-tuned this model by adding a classification layer on top of the mBERT model, following the recommendation of the original work \cite{devlin-etal-2019-bert}.
    \item \textbf{XLM-R}: This model was introduced in the work of \cite{xlm-r} for cross-lingual language understanding tasks. It was trained on a multilingual training dataset with 137 GB of Vietnamese text. In this work, we only employ the base version for our experiments.
    \item \textbf{PhoBERT}: PhoBERT is one of the first public encoder-type language models pre-trained for monolingual Vietnamese. The initial model was pre-trained on 20GB of uncompressed text data, while the second model was trained on a combination of 20GB of Wikipedia and news texts along with 120GB of texts from OSCAR-2301. 
\end{itemize}

\textbf{Comparative Element Extraction Task}: We fine-tune the viT5 and mT5 models specifically for a single task without combining prompt instructions. This implies that we only feed the input sentence to the model to get the prediction.
\begin{itemize}
    \item \textbf{mT5}: mT5 \cite{mt5} is a multilingual variant of T5 which was trained on the MC4 dataset covering 101 languages, including the Vietnamese. The previous work \cite{thin_finetuning} demonstrated that fine-tuning mT5 also gives comparative results for different types of classification problems.
    \item \textbf{viT5}: This model \cite{vit5}, represents a monolingual variant of the T5 architecture tailored for the Vietnamese language. viT5 was trained in two versions, namely the base and large models. We fine-tuned both versions using the text-to-text format for the CSI task. 
\end{itemize}

\subsection{Main results}
\begin{table}[t]
\caption{\centering The experimental results for the Comparative Sentence Identification sub-task on the public test.}
\label{result_CSI}
\centering
\resizebox{0.5\textwidth}{!}{%
\begin{tabular}{llccc}
\hline
\textbf{Type} & \textbf{Model} & \textbf{Precision} & \textbf{Recall} & \textbf{F1-score} \\ \hline
\multirow{4}{*}{Encoder-only} 
 & mBERT & 0.742 & 0.8653 & 0.7989 \\
 & XLM-R & 0.7429 & 0.9026 & 0.8150 \\
 & PhoBERTv1 & 0.7435 & 0.8968 & 0.8130 \\
 & PhoBERTv2 &  \textbf{0.7606} & \textbf{0.9284} & \textbf{0.8361} \\ \hline
\end{tabular}%
}
\end{table}

\begin{table*}[t]
\caption{\centering Results of different approaches for Comparative Opinion Quintuplet Extraction on the private test are presented. We use PhoBERTv2 for identifying comparative reviews. }
\label{result_COQE}
\centering
\resizebox{\textwidth}{!}{%
\begin{tabular}{llllllll}
\hline
Type & \textbf{Model} & \multicolumn{1}{c}{\textbf{\begin{tabular}[c]{@{}c@{}}Macro \\ Precision\end{tabular}}} & \multicolumn{1}{c}{\textbf{\begin{tabular}[c]{@{}c@{}}Macro \\ Recall\end{tabular}}} & \multicolumn{1}{c}{\textbf{\begin{tabular}[c]{@{}c@{}}Macro \\ F1-score\end{tabular}}} & \multicolumn{1}{c}{\textbf{\begin{tabular}[c]{@{}c@{}}Micro \\ Precision\end{tabular}}} & \multicolumn{1}{c}{\textbf{\begin{tabular}[c]{@{}c@{}}Micro\\ Recall\end{tabular}}} & \multicolumn{1}{c}{\textbf{\begin{tabular}[c]{@{}c@{}}Micro \\ F1-score\end{tabular}}} \\ \hline
\multirow{4}{*}{Baselines} 
 & mT5-base & 0.0810 & 0.0730 & 0.0770 & 0.1667 & 0.1509 & 0.1584 \\
 & mT5-large & 0.1419 & 0.1411 & 0.1409 & 0.2254 & 0.2225 & 0.2239 \\
 & viT5-base & 0.1403 & 0.1037 & 0.1190 & 0.2416 & 0.1751 & 0.2031 \\
 & viT5-large & 0.2055 & 0.1563 & 0.1764 & 0.2758 & 0.2181 & 0.2435 \\ \hline

\multirow{4}{*}{Baselines with Data Augmentation} 
 & mT5-base & 0.1089 & 0.1059 & 0.1073 & 0.1948 & 0.1883 & 0.1915 \\
 & mT5-large & 0.1319 & 0.1371 & 0.1341  & 0.2308 & 0.2313 & 0.2310 \\
 & viT5-base & 0.1204 & 0.0875 & 0.1012 & 0.2088 & 0.1509 & 0.1752 \\
 & viT5-large & 0.2111 & 0.1610 & 0.1817 & 0.2796 & 0.2236 & 0.2485 \\ \hline
 
\multirow{4}{*}{\begin{tabular}[c]{@{}l@{}}Multitask Instruction Tuning\end{tabular}} 
 & mT5-base & 0.1206 & 0.1212 & 0.1205 & 0.2230 & 0.2203 & 0.2216 \\
 & mT5-large & 0.2465 & 0.1511 & 0.1654 & 0.2332 & 0.2368 & 0.2350 \\
 & viT5-base & 0.1549 & 0.1312 & 0.1416 & 0.2773 & 0.2269 & 0.2495 \\
 & viT5-large & 0.2417 & 0.2180 & 0.2251 & 0.2813  & 0.2941 & 0.2876  \\ \hline
\multirow{4}{*}{\begin{tabular}[c]{@{}l@{}}Multitask Instruction Tuning with \\ Data Augmentation\end{tabular}} 
 & mT5-base & 0.1259 & 0.1303 & 0.1279 & 0.2040 & 0.2004 & 0.2022 \\
 & mT5-large & 0.2158 & 0.1777 & 0.1934 & 0.2873 & \textbf{0.3166} & 0.2629 \\
 & viT5-base & 0.1936 & 0.1502 & 0.1682 & 0.2699 & 0.2203 & 0.2460 \\
 & viT5-large & \textbf{0.2862} & \textbf{0.2216} & \textbf{0.2373}  & \textbf{0.2880} & 0.3029 & \textbf{0.2952} \\ \hline
\end{tabular}%
}
\end{table*}

\begin{table*}[t]
\caption{\centering The results of our best model are compared with the five top participants on the private scoreboard (Best scores are in the bold).}
\label{result_competion}
\centering
\resizebox{\textwidth}{!}{%
\begin{tabular}{ll|ccc|ccc}
\hline
\textbf{Rank} & \multicolumn{1}{l|}{\textbf{Team}} & \textbf{\begin{tabular}[c]{@{}c@{}}Macro \\ Precision\end{tabular}} & \textbf{\begin{tabular}[c]{@{}c@{}}Macro\\ Recall\end{tabular}} & \textbf{\begin{tabular}[c]{@{}c@{}}Macro \\ F1-score\end{tabular}} & \textbf{\begin{tabular}[c]{@{}c@{}}Micro \\ Precision\end{tabular}} & \textbf{\begin{tabular}[c]{@{}c@{}}Micro\\ Recall\end{tabular}} & \textbf{\begin{tabular}[c]{@{}c@{}}Micro\\ F1-score\end{tabular}} \\ \hline
1 & \textbf{thindang (Ours)} & \textbf{0.2862} & 0.2216 & \textbf{0.2373} & 0.2880 & 0.3029 & \textbf{0.2952} \\ \hline
2 & pthutrang513 & 0.2021 & \textbf{0.2718} & 0.2300 & 0.2234 & \textbf{0.3359} & 0.2684 \\ \hline
3 & thanhlt998 & 0.2093 & 0.2199 & 0.2131 & \textbf{0.2941} & 0.2941 & 0.2941 \\ \hline
4 & duyvu1110 & 0.0964 & 0.1375 & 0.1119 & 0.1709 & 0.2698 & 0.2092 \\ \hline
5 & ComOM\_RTX5000 & 0.0968 & 0.1065 & 0.0997 & 0.1675 & 0.1894 & 0.1778 \\ \hline
\end{tabular}%
}
\end{table*}

Table \ref{result_CSI} presents the experimental results for the Comparative Sentence Identification task, while Table \ref{result_COQE} displays the results of different models for the Comparative Opinion Quintuple Extraction (COQE) task. For the CSI task, we report the Precision, Recall, and binary F1-score. As shown in Table \ref{result_CSI}, it can be observed that the PhoBERTv2 model outperforms the three remaining models, including PhoBERTv1, mBERT, and XLM-R. The reason this model achieved the best results is its utilization as a pre-trained monolingual language model, which was trained on a combination of 20GB of Vietnamese Wikipedia and News texts along with 120GB of the OSCAR-2301 corpus.

For the COQE task, we utilize the micro-average and macro-average of Precision, Recall, and F1-score under an exact match of the quintuplets to compare all models. Results of different baselines and our solutions are reported in Table \ref{result_COQE}. Experimental results show that our multi-task instruction tuning with data augmentation technique achieves the best performance in both macro and micro F1-score. In Table \ref{result_COQE}, the monolingual viT5 model yields better results than the multilingual version of T5 in both cases. Additionally, it is worth noting that all of our proposed components (data augmentation and multi-task instruction tuning) in this work improve the results compared to the baseline models. Specifically, the data augmentation technique improves the performance of the individual baseline models. When compared to the baseline models, our multi-task instruction tuning strategy significantly enhances their performance. This result demonstrates that fine-tuning generative models with the instruction template and multi-task strategy helps the models learn the correlation information between elements in the quintuplet. It is noteworthy that our multi-task instruction tuning, coupled with data augmentation based on the viT5-large model, achieves the highest performance in terms of the two F1 scores on the private test. This result helps our final submission achieve the top position on the shared-task scoreboard (as shown in Table \ref{result_competion}). Our models consistently yield the best results in both F1-score measurements under exact match.

\section{Conclusion}
\label{conclusion}
In this study, we introduce our participant system for the cOMom at VLSP2023 shared task. The purpose of this task is to identify comparative reviews and then extract all comparative quintuples mentioned. To address this task, we propose a two-stage system based on fine-tuning a BERTology model for the CSI task and unified multi-task instruction tuning for the CEE task. Besides, we apply the simple data augmentation technique to increase the size of the dataset for training our model in the second stage. Experimental results show that our approach outperforms the other competitors and has achieved the top score on the official private test. In the future, we plan to exploit the fine-tuning of some large language models such as BloomZ \cite{bloomZ}, LLaMa-2 \cite{touvron2023llama}, or PhoGPT \cite{PhoGPT} in order to achieve robust performance.

\section*{Acknowledgments}

The authors would like to express their profound gratitude to the Vietnamese Language and Speech Processing (VLSP) organisers for their exceptional effort in coordinating the Comparative Opinion Mining from Vietnamese Product Reviews challenge. Dang Van Thin was funded by the Master, PhD Scholarship Programme of Vingroup Innovation Foundation (VINIF), code VINIF.2023.TS.117. 

\bibliography{vlsp}
\bibliographystyle{aclnatbib}
\end{document}